\documentclass[runningheads]{llncs}

\usepackage{eccv}

\usepackage{eccvabbrv}

\usepackage{graphicx}
\usepackage{booktabs}
\usepackage{multirow}
\usepackage[table]{xcolor}
\usepackage[accsupp]{axessibility}  

\usepackage[pagebackref]{hyperref}

\usepackage{orcidlink}

\begin{document}

\title{MS-Resampler: Multi-Scope Visual Resampling for Efficient Multimodal LLMs} 

\titlerunning{MS-Resampler}

\author{Zhongyang Li\inst{1,2} \and
Yaqian Li\inst{1} \and 
Faming Fang\inst{2}\textsuperscript{*}\and 
Rinyoichi Takezoe\inst{1} \and 
Zi-Hao Bo\inst{1} \and 
Cheng Qian\inst{1} \and 
Mo Guang\inst{1} \and 
Guixu Zhang\inst{2} \and
Kaiwen Long\inst{1}\textsuperscript{*}
}


\institute{Li Auto Inc. \and East China Normal University.}

\maketitle
\begingroup\renewcommand{\thefootnote}{*}\footnotetext{Corresponding authors: Faming Fang and Kaiwen Long.}\endgroup

\begin{abstract}
Multimodal large language models (MLLMs) typically employ resampling-based projectors to transform dense visual features into a compact token sequence for language modeling. Most existing resamplers adopt a single, fixed aggregation scope via global cross-attention, which can blur fine-grained local evidence and limit the ability to capture both local details and global context within a fixed token budget. 
In this work, we propose MS-Resampler, a multi-scope visual resampling framework for MLLMs. MS-Resampler instantiates multiple scope-specific resamplers by injecting explicit spatial scope priors into the resampling attention, enabling each branch to aggregate visual information at a particular granularity from local to global. The outputs of these scope-specific resamplers are then adaptively fused to produce the final visual representations for language modeling.
Extensive experiments on ten public multimodal benchmarks show that MS-Resampler consistently improves visual understanding and multimodal reasoning over conventional single-scope resamplers, while introducing only minimal computational overhead.
\keywords{Multimodal Large Language Models \and Visual Projector \and Visual Token Reduction \and Multi-Scale Representation}
\end{abstract}
\begin{figure}[t!]
  \centering
  \includegraphics[width=\linewidth]{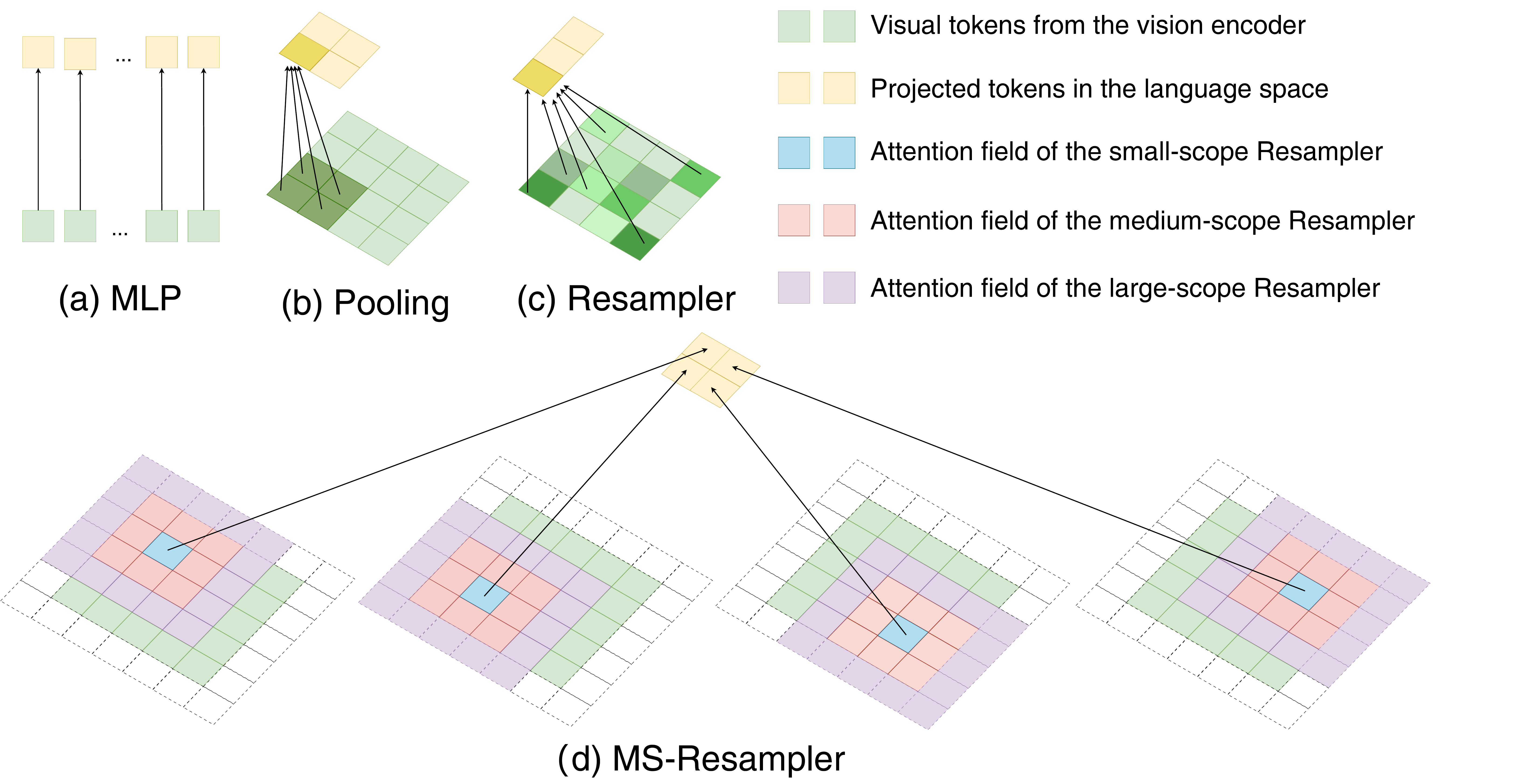}
  \caption{
  Comparison of visual projectors in MLLMs. (a) MLP maps patch tokens independently, leading to redundant inputs. (b) Global resampler aggregates with a single global scope, which may dilute local evidence. (c) MS-Resampler uses multiple scoped resamplers and fuses their outputs to capture local-to-global semantics under a fixed token budget.
}
  \label{fig_1}
\end{figure}
\section{Introduction}

Multimodal large language models (MLLMs) extend large language models with visual inputs, enabling unified understanding and reasoning over images and text. Most modern MLLMs follow a frozen-backbone paradigm: a pretrained vision encoder~\cite{clip,siglip} extracts dense visual features, and a language model~\cite{gpt4,qwen} performs sequence modeling and generation. In this pipeline, the visual projector converts high-dimensional visual features into a compact sequence of tokens compatible with the language model. Its design is crucial for both efficiency and performance, as it determines the visual token budget and the extent to which visual evidence is preserved for cross-modal reasoning.

Early MLLMs commonly adopt MLP-based projectors, as exemplified by LLaVA-style models~\cite{llava,llava15}. These projectors perform per-token channel mapping from patch features to the language embedding space while keeping the token count unchanged. As image resolutions increase and multi-image or video inputs become more common, the visual sequence length grows rapidly, making redundancy a practical bottleneck. To reduce the token budget more explicitly, a line of work compresses visual tokens through local merging. Representative designs aggregate neighboring patches with fixed or learnable local operators, including coarse-to-fine pooling (e.g., TokenPacker)~\cite{tp}, lightweight locality-injected projections with depth-wise convolutions (e.g., LDP-v2)~\cite{ldp}, and locality-preserving abstraction modules (e.g., C-/D-Abstractor)~\cite{cabs}. In parallel, resampling-based projectors~\cite{blip2,QwenVL,MQT} introduce a fixed set of learnable query tokens and use cross-attention to distill dense visual tokens into a compact latent set. This query-based resampling provides a flexible bottleneck for controlling the token budget and has become a widely adopted design in modern MLLMs.

Despite steady progress, existing projector designs still face fundamental limitations under a fixed token budget, as illustrated in Fig.~\ref{fig_1}. MLP projectors preserve a long visual sequence, which retains substantial redundancy and provides little structural inductive bias for spatial aggregation. Local merging methods are efficient and stable, but they typically operate at a single granularity determined by the kernel, stride, or merging rule. This single-scale aggregation may sacrifice either fine-grained evidence or broader contextual relations, depending on the chosen scale. Resampling-based projectors offer greater modeling capacity through attention-based aggregation, yet standard designs use global cross-attention for all queries. This design can bias representations toward coarse scene-level patterns and weaken localized evidence such as small text, subtle attributes, or occluded parts.

We argue that this single-scope assumption is a key limitation for visual-language alignment. Visual semantics inherently span multiple spatial scopes. Fine-grained concepts depend on local neighborhoods, whereas compositional reasoning and scene understanding require broader context and long-range relations. Under a fixed token budget, a single scope imposes an implicit trade-off between local fidelity and global context. It can dilute one type of information in favor of the other.

To address this limitation, we propose \textbf{MS-Resampler}, a multi-scope visual projector that captures hierarchical visual semantics by combining multiple scope-specific resampling branches (Fig.~\ref{fig_1}d). Our core building block is the \textbf{Scoped Resampler}, which augments standard resampling attention with an explicit scope prior implemented as a structured spatial attention bias. This design controls the effective receptive field of aggregation. MS-Resampler instantiates $E$ parallel scoped resamplers spanning local-to-global receptive fields and fuses their outputs at the token level to produce compact visual tokens. This plug-and-play architecture remains simple while enabling both fine-grained evidence and global context to be encoded within the same token budget.

Our contributions are summarized as follows:
\begin{itemize}
    \item We introduce \textbf{Scoped Resampler}, a scope-controlled resampling module that injects explicit spatial scope priors into cross-attention, enabling aggregation at a specified visual granularity.
    \item We propose \textbf{MS-Resampler}, which combines multiple scoped resamplers spanning local-to-global receptive fields and fuses their outputs to encode hierarchical visual semantics under a fixed token budget.
    \item Extensive experiments on ten public multimodal benchmarks demonstrate that \textbf{MS-Resampler} consistently improves visual understanding and multimodal reasoning over conventional single-scope projectors with minimal computational overhead.
\end{itemize}
\section{Related Work}
\subsection{Multimodal Large Language Models (MLLMs)}
Multimodal large language models (MLLMs) extend large language models with visual inputs, enabling unified reasoning and generation across vision and language. Early vision-language pretraining frameworks such as CLIP~\cite{clip} and ALIGN~\cite{align} learn transferable aligned representations from large-scale image-text pairs and have become standard visual backbones for subsequent MLLMs. Building on these representations, models such as Flamingo~\cite{flamingo}, BLIP-2~\cite{blip2}, and the LLaVA family~\cite{llava,llava15} integrate strong vision encoders (e.g., CLIP-ViT and SigLIP-ViT~\cite{clip,siglip}) with capable LLMs (e.g., LLaMA~\cite{llama} and Phi~\cite{phi2}) to support a wide range of multimodal tasks. Despite architectural variations, a common pipeline emerges: the vision encoder produces dense visual tokens, which are transformed by a visual projector into representations compatible with the LLM. Under the frozen-backbone paradigm, the projector becomes the key modality bridge, as it determines what visual information is conveyed to the LLM and at what token budget.

\subsection{Strategies for Visual Token Reduction in MLLMs}
Modern vision encoders output dense token sequences, which substantially increase computation and memory in the LLM. Existing work reduces this overhead at two stages: compressing tokens before entering the LLM and dynamically reducing tokens inside the LLM.

\textbf{Projector-Level Compression.}  
A straightforward approach is to reduce visual tokens before they enter the LLM. Early MLLMs such as LLaVA‑1.5~\cite{llava15} use MLP that maintain one-to-one mappings from image patches to language-compatible embeddings, preserving significant redundancy. Later methods focus on structural compression and feature condensation. TokenPacker~\cite{tp} aggregates tokens using a coarse-to-fine strategy, while LDP-v2~\cite{ldp} leverages lightweight depth-wise convolutions to inject local spatial priors. C‑Abstractor~\cite{cabs} improves locality preservation using convolutional positional abstractions. LLaVA Mini~\cite{llavamini} explores extreme compression, reducing an entire image to very few tokens with modality pre-fusion to maintain performance. Recent advances include SAEP~\cite{SAEP} for multi-layer spatial token aggregation, Delta-LLaVA~\cite{delta-llava} using base-then-specialize alignment, and VisionSelector~\cite{visionselector} for learnable token selection.

\textbf{In-LLM Visual Token Reduction.}  
Beyond the projector, dynamic strategies reduce token redundancy within the transformer layers of the LLM. Progressive reduction frameworks introduce hollow attention and dynamic FFN activation~\cite{beyond}, search-based methods optimize per-layer token budgets~\cite{accelerating}, and diversity-driven pruning or merging schemes such as DivPrune balance information retention with token reduction~\cite{divprune}. These methods are effective for lowering computation, but they typically operate on a single aggregation perspective without explicitly modeling hierarchical or multi-scale visual semantics.

Overall, existing strategies primarily target the \emph{quantity} of visual tokens or the cost of self-attention, leaving open the challenge of designing compact visual representations that are semantically expressive across multiple spatial scales.

\subsection{Multi-Scale Representation}
Multi-scale representations are a long-standing theme in computer vision, motivated by the need to capture both local details and global structure. Hierarchical backbones progressively build multi-level features by aggregating fine-grained information into coarser representations, as exemplified by Swin Transformer~\cite{swin}. Multi-path or multi-resolution designs such as MPViT~\cite{mpvit} fuse features across scales to improve representational diversity, while MaxViT~\cite{maxvit} combines local and global interactions through multi-axis attention. Scale-aware Transformers also introduce explicit mechanisms to enlarge or diversify receptive fields, such as ScopeViT~\cite{ScopeViT}. These results suggest that explicitly modeling multiple spatial scopes can enrich representations, which is particularly relevant to MLLMs.

\begin{figure*}[t!]
  \centering
  \includegraphics[width=\linewidth]{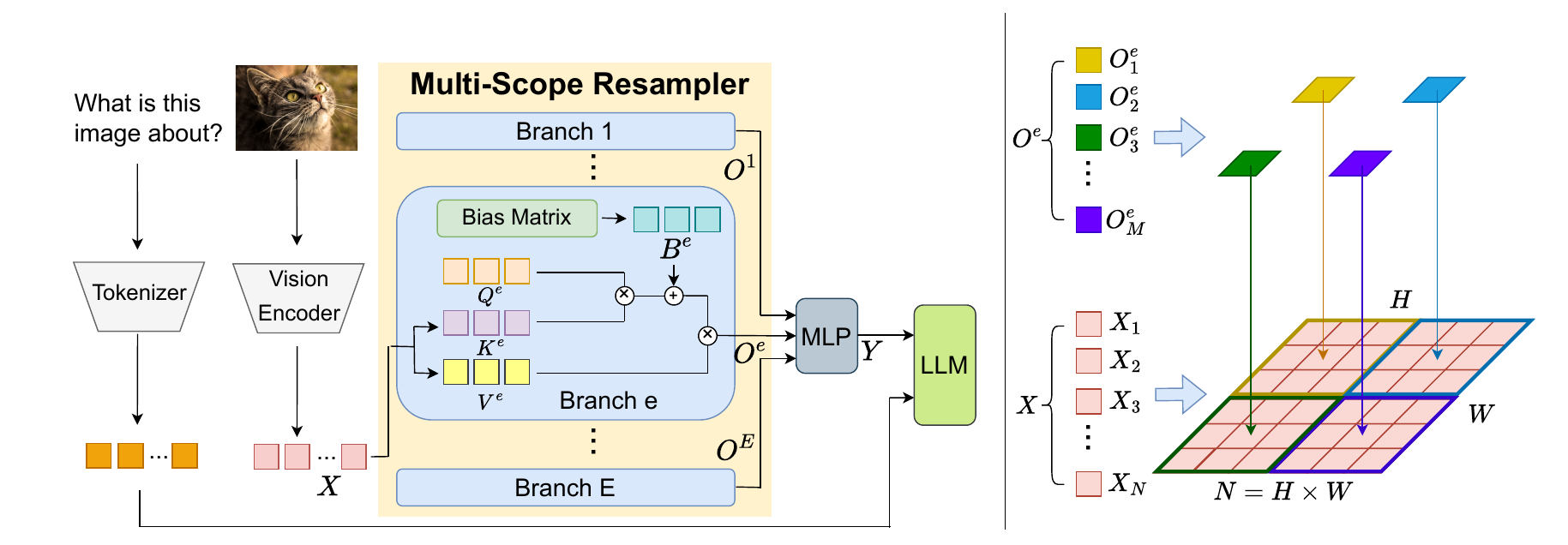}
  \caption{
  Overview of MS-Resampler. 
Dense visual tokens extracted by the vision encoder are processed by multiple scoped resampling branches, each operating with a different spatial aggregation scope. 
The outputs from these branches are fused to produce the final compact visual tokens for the language model.
}
  \label{fig_2}
\end{figure*}
\section{Method}
\subsection{Preliminaries}

In multimodal large language models (MLLMs), a vision encoder produces a sequence of visual tokens $\mathbf{X}\in\mathbb{R}^{N\times d}$, where $N$ denotes the number of tokens and $d$ the feature dimension. 
A visual projector maps these tokens to $\mathbf{Y}\in\mathbb{R}^{M\times c}$ with $M\ll N$, where $c$ matches the language embedding dimension. 
The resulting tokens $\mathbf{Y}$ are then fed into the language model as visual conditioning for subsequent multimodal reasoning and generation.

To improve the quality of compressed visual representations, we propose \textbf{MS-Resampler}, a multi-scope visual projection module (Fig.~\ref{fig_2}). 
Given dense visual features from a frozen vision encoder, MS-Resampler constructs $E$ parallel \emph{scoped resampling} branches. 
Each branch aggregates visual tokens within a predefined spatial scope, enabling the model to capture complementary semantics across different spatial granularities, ranging from fine-grained local details to global contextual context. 
The outputs of these branches are then fused in the embedding space to produce the final visual tokens, which are subsequently passed to the language model.

\subsection{Vanilla Resampling Projector}
As a common choice of visual projector in modern MLLMs, resampling-based projectors use a fixed set of learnable queries to distill dense visual tokens into a compact sequence via global cross-attention. 
Given visual tokens $\mathbf{X}\in\mathbb{R}^{N\times d}$, we introduce learnable query embeddings $\mathbf{Q}\in\mathbb{R}^{M\times c}$ and linearly project $\mathbf{X}$ into keys and values:
\begin{equation}
\mathbf{K}=\mathbf{X}\mathbf{W}_K\in\mathbb{R}^{N\times c},\quad
\mathbf{V}=\mathbf{X}\mathbf{W}_V\in\mathbb{R}^{N\times c},
\end{equation}
where $\mathbf{W}_K$ and $\mathbf{W}_V$ are learnable projection matrices. 
The resampled tokens are then computed as
\begin{equation}
\mathbf{Y}=
\mathrm{Softmax}\left(
\frac{\mathbf{Q}\mathbf{K}^\top}{\sqrt{c}}
\right)\mathbf{V},
\label{eq:vanilla_resampler}
\end{equation}
where $\mathbf{Y}\in\mathbb{R}^{M\times c}$ and typically $M\ll N$. 
This formulation performs global aggregation: every query attends to all $N$ visual tokens, resulting in a single shared receptive scope for resampling.

\subsection{Scoped Resampling Branch}
\label{subsec:scoped_resampler}

The vanilla resampling projector performs global cross-attention, where each learnable query attends to all visual tokens. 
This global aggregation is effective for capturing holistic semantics, but it assumes that visual evidence can be integrated under a single receptive scope. 
However, visual concepts span multiple spatial granularities. Under a fixed token budget, global attention may over-emphasize dominant scene-level patterns and under-represent localized evidence such as small text, small objects, or subtle attributes.

To explicitly control the aggregation scope during resampling, we introduce a \textbf{Scoped Resampler} branch. 
It enables \emph{scope-controlled} resampling by imposing an explicit spatial scope prior on the cross-attention distribution, implemented as a structured additive term on the attention logits. 
For the $e$-th branch, we use learnable queries $\mathbf{Q}^{(e)}\in\mathbb{R}^{M\times c}$ and project visual tokens as
\[
\mathbf{K}^{(e)}=\mathbf{X}\mathbf{W}^{(e)}_K\in\mathbb{R}^{N\times c},\qquad
\mathbf{V}^{(e)}=\mathbf{X}\mathbf{W}^{(e)}_V\in\mathbb{R}^{N\times c},
\]
where $\mathbf{W}^{(e)}_K$ and $\mathbf{W}^{(e)}_V$ are learnable projection matrices. 
The scoped resampling output is computed as
\begin{equation}
\mathbf{O}^{(e)}=
\mathrm{Softmax}\left(
\frac{\mathbf{Q}^{(e)}(\mathbf{K}^{(e)})^\top}{\sqrt{c}}
+\beta\,\mathbf{B}^{(e)}
\right)\mathbf{V}^{(e)},
\label{eq:scoped_attn}
\end{equation}
where $\mathbf{O}^{(e)}\in\mathbb{R}^{M\times c}$ is the output of branch $e$, $\mathbf{B}^{(e)}\in\mathbb{R}^{M\times N}$ is a spatial attention bias matrix, and $\beta$ controls the strength of the scope prior.

The additive term in Eq.~(\ref{eq:scoped_attn}) biases the attention logits and thus differs from a hard mask. 
It does not remove connections outside the preferred region; instead, it down-weights them relative to in-scope tokens, while still allowing cross-scope aggregation when needed. 
When $\beta=0$, Eq.~(\ref{eq:scoped_attn}) reduces to the vanilla global resampler in Eq.~(\ref{eq:vanilla_resampler}).

\begin{figure*}[t!]
  \centering
  \includegraphics[width=\linewidth]{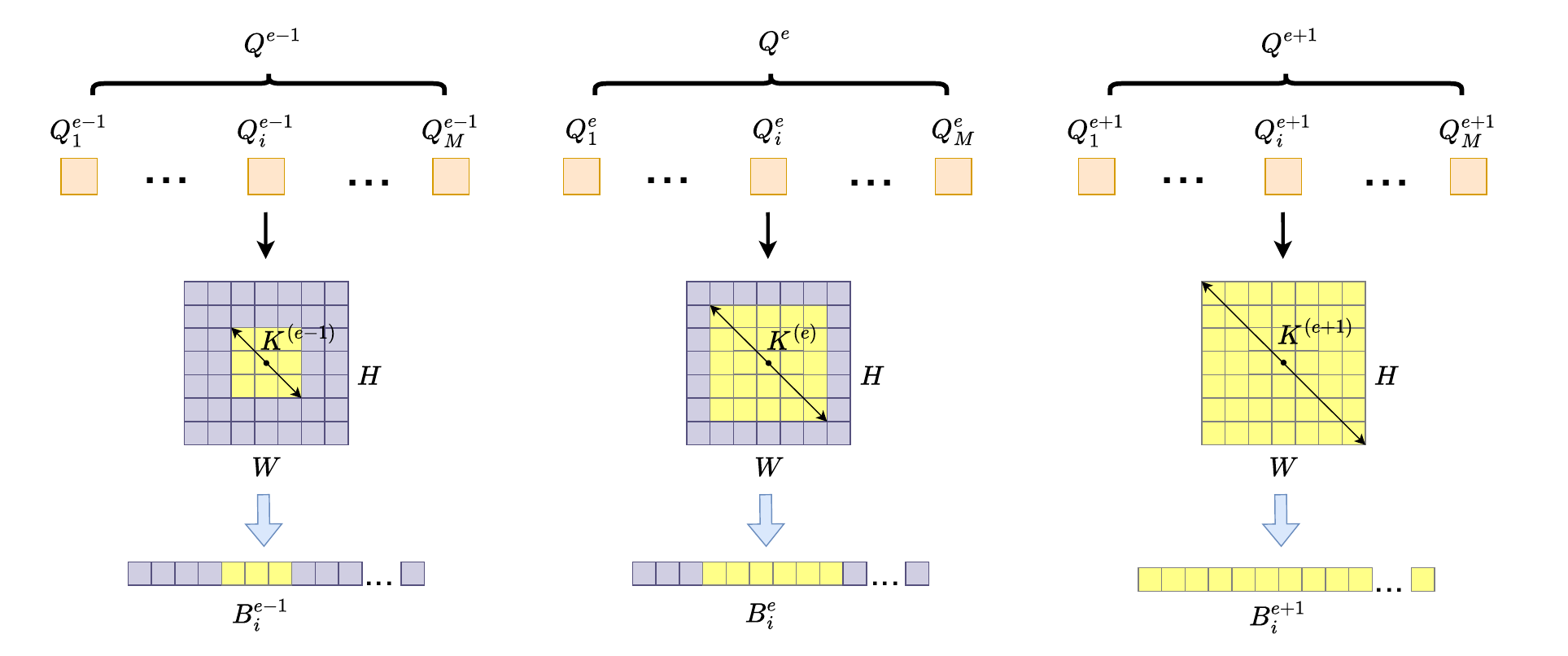}
  \caption{
    Illustration of spatial attention bias construction for scoped resampling. Each query corresponds to an anchor position on the visual token grid. For branch $e$, a window of size $k^{(e)}$ defines its preferred spatial scope. 
}
  \label{fig_3}
\end{figure*}
\subsection{Spatial Attention Bias Construction}
\label{subsec:scope_construction}

Each scoped resampling branch encodes its preferred aggregation scope through a spatial attention bias matrix $\mathbf{B}^{(e)}\in\mathbb{R}^{M\times N}$. 
This matrix specifies, for each query, which spatial locations on the visual token grid should be prioritized during resampling (Fig.~\ref{fig_3}). 
Below, we describe how $\mathbf{B}^{(e)}$ is constructed from query anchors and a branch-specific window size.

We first reshape the visual token sequence into a 2D grid of size $H\times W$, where $N=H\cdot W$, and denote the coordinate of the $l$-th token as $(y_l,x_l)$. 
To assign spatial responsibilities to the $M$ learnable queries, we associate each query with an anchor location $\{(y_i,x_i)\}_{i=1}^{M}$ on the grid. 
In practice, we partition the $H\times W$ grid into $M$ non-overlapping regions and choose the center of each region as its anchor, which yields a uniform spatial coverage.

Given the anchor of query $i$, branch $e$ defines its preferred aggregation neighborhood by a window size $k^{(e)}$. 
Specifically, we define the neighborhood as a $k^{(e)}\times k^{(e)}$ window centered at $(y_i,x_i)$:
\begin{equation}
\Omega_i^{(e)}=
\left\{
(y,x)\ \middle|\ 
|y-y_i|\le \left\lfloor k^{(e)}/2\right\rfloor,\ 
|x-x_i|\le \left\lfloor k^{(e)}/2\right\rfloor
\right\}.
\label{eq:scope_def}
\end{equation}
The window size $k^{(e)}$ directly controls the effective aggregation scope of branch $e$.

Finally, we initialize the bias matrix $\mathbf{B}^{(e)}$ by assigning higher preference to tokens within $\Omega_i^{(e)}$:
\begin{equation}
B^{(e)}_{i,l}=
\begin{cases}
0, & \text{if } (y_l,x_l)\in\Omega_i^{(e)},\\
-1, & \text{otherwise}.
\end{cases}
\label{eq:bias_init}
\end{equation}
This bias is added to the attention logits in Eq.~(\ref{eq:scoped_attn}) as a soft scope prior: in-window tokens are favored, whereas out-of-window tokens are down-weighted rather than masked out. 
This design preserves the ability to attend beyond the preferred neighborhood when cross-scope aggregation is beneficial. 
By instantiating different branches with different $k^{(e)}$, we obtain scoped resamplers with complementary aggregation ranges.

\subsection{Multi-Scope Resampling and Fusion}
\label{subsec:multi_scope_fusion}

Based on the scoped resampling formulation in Eq.~(\ref{eq:scoped_attn}) and the bias construction in Sec.~\ref{subsec:scope_construction}, we instantiate $E$ scoped resampling branches with different window sizes $\{k^{(1)},\dots,k^{(E)}\}$. For the same visual token sequence $\mathbf{X}$, each branch produces a compact set of resampled tokens $\mathbf{O}^{(e)}\in\mathbb{R}^{M\times c}$:
\begin{equation}
\mathcal{O}=\{\mathbf{O}^{(1)},\mathbf{O}^{(2)},\dots,\mathbf{O}^{(E)}\}.
\label{eq:outputs_set}
\end{equation}
Different window sizes yield complementary aggregation scopes: smaller $k^{(e)}$ prioritizes local fine-grained evidence, whereas larger $k^{(e)}$ captures broader context and long-range relations. When the window covers the entire grid, $\mathbf{B}^{(e)}$ degenerates to an all-zero matrix; similarly, setting $\beta=0$ removes the scope prior and reduces Eq.~(\ref{eq:scoped_attn}) to the vanilla global resampler.

To integrate multi-scope information without increasing the LLM input length, we keep the output token count fixed to $M$ and perform \emph{position-wise fusion}. Since all branches share the same set of query anchors, the $i$-th output token in each branch corresponds to the same spatial reference location. We therefore fuse branch outputs at the same token index $i$ to combine evidence gathered under different scopes. Concretely, for each token index $i\in\{1,\dots,M\}$ and channel $j\in\{1,\dots,c\}$, we form an expert vector
\begin{equation}
\mathbf{z}_{i,j}=
\big[\mathbf{O}^{(1)}_{i,j},\mathbf{O}^{(2)}_{i,j},\dots,\mathbf{O}^{(E)}_{i,j}\big]^{\top}
\in\mathbb{R}^{E},
\label{eq:expert_vec}
\end{equation}
and fuse it using a two-layer MLP shared across all $(i,j)$:
\begin{equation}
\mathbf{O}_{\mathrm{final},\,i,j}
=
\mathbf{w}_2^{\top}\,
\phi\!\left(\mathbf{W}_1\mathbf{z}_{i,j}+\mathbf{b}_1\right)+b_2,
\label{eq:expert_fusion}
\end{equation}
where $\mathbf{W}_1\in\mathbb{R}^{h\times E}$, $\mathbf{b}_1\in\mathbb{R}^{h}$, $\mathbf{w}_2\in\mathbb{R}^{h}$, and $b_2\in\mathbb{R}$ are learnable parameters, $h$ is the hidden dimension, and $\phi(\cdot)$ is a non-linear activation. Sharing fusion parameters across positions ensures that fusion operates only along the scope (expert) dimension and does not mix information across different token indices. The fused tokens $\mathbf{O}_{\mathrm{final}}\in\mathbb{R}^{M\times c}$ serve as the final projected visual representation and are fed into the LLM.
\begin{table*}[t!]
  \centering
  \caption{Performance comparison with pre-LLM compression methods on LLaVA-1.5. The best and second-best results are highlighted in \textbf{bold} and \underline{underlined}.}
  \setlength{\tabcolsep}{1pt} 
  {
  \fontsize{7pt}{9pt}\selectfont
  \begin{tabular}{l | c c c c c c c c c c | c}
    \hline\hline
    Methods & VQA\textsuperscript{T} & POPE & VQA\textsuperscript{v2} & SQA & MM-Vet & 
    GQA & MME\textsuperscript{P} & MMB & Seed\textsuperscript{I} & MMStar & Avg.\\
    \hline

    \rowcolor{gray!10}
    \multicolumn{12}{c}{ LLaVA-1.5-7B, Vision token = 144}\\
    \hline
    Vanilla  & 58.2 & 85.9 & 78.5 & 68.0 & 31.6 & 62.0 & 1511 & 64.3 & 66.1 & 31.9 & 100.00\%\\
    Pixel-Shuffle~\cite{InternVL15} & 54.2 & \underline{87.0} & 75.6 & 69.6 & 29.1 & 60.2 & 1422 & 64.5 & 61.6 & 31.7 & 96.92\% \\
    FasterVLM~\cite{FasterVLM}  & \textbf{57.2} & 83.5 & 76.2 & 67.8 & 32.2 & 58.0 & 1440 & 64.3 & 62.3 & 31.3 & 97.54\%\\
    C-Abstractor~\cite{cabs} & 54.1 & 85.9 & 75.7 & 68.5 & 28.2 & 60.4 & 1392 & 63.3 & 62.3 & 31.1 & 95.91\%\\
    MQT-LLaVA~\cite{MQT}  & 52.6 & 86.2 & 76.8 & 67.6 & 29.8 & 61.4 & 1445 & 64.4 & 62.0 & 31.9 & 97.09\% \\
    TokenPacker~\cite{tp} & \underline{57.1} & \underline{87.0} & \textbf{78.0} & 68.9 & \underline{33.0} & \textbf{62.0} & \textbf{1454} & 65.0 & \underline{65.1} & \underline{32.6} & \underline{100.25\%} \\
    LDP-v2~\cite{ldp}    & 56.4 & 86.3 & 77.3 & \underline{70.9} & 31.7 & 61.5 & 1435 & \underline{65.1} & 63.9 & 31.9 & 99.25\%\\
    SAEP~\cite{SAEP} & 56.4 & \textbf{87.2} & \underline{77.6} & 69.0 & 32.8 & \underline{61.8} & 1424 & 63.8 & 64.8 & 31.7 & 99.31\% \\
    \textbf{MS-Resampler}& \textbf{57.2} & \underline{87.0} & \textbf{78.0} & \textbf{71.2} & \textbf{33.1} & \textbf{62.0} & \underline{1453} & \textbf{65.6} & \textbf{65.2} & \textbf{33.9} & \textbf{101.15\%}\\
    \hline

    \rowcolor{gray!10}
    \multicolumn{12}{c}{ LLaVA-1.5-7B, Vision token = 64}\\
    \hline
    Pixel-Shuffle~\cite{InternVL15} & 52.2 & 85.5 & 73.2 & 68.7 & 25.8 & 58.7 & 1383 & 64.4 & 59.3 & 32.1 & 94.18\% \\
    FasterVLM~\cite{FasterVLM} & \underline{56.0} & 80.4 & 72.5 & 68.8 & 28.7 & 55.0 & 1342 & 61.6 & 57.9 & 31.8 & 93.48\% \\
    C-Abstractor~\cite{cabs}  & 51.3 & 85.7 & 73.7 & 68.6 & 26.1 & 59.1 & 1415 & 61.8 & 60.0 & 32.3 & 94.24\% \\
    MQT-LLaVA~\cite{MQT} & 51.6 & 83.6 & 75.3 & 67.0 & 28.9 & 60.0 & \underline{1454} & 63.5 & 60.8 & 30.9 & 95.25\% \\
    TokenPacker~\cite{tp} & 55.6 & 86.5 & \underline{77.1} & 69.4 & 29.1 & \underline{61.0} & 1440 & 64.4 & 63.0 & 31.3 & 97.59\% \\
    SAEP~\cite{SAEP} & 53.9 & 86.0 & 75.6 & 68.0 & \underline{30.5} & 60.3 & 1380 & 62.9 & \underline{63.4} & \underline{33.1} & 97.16\% \\
    LDP-v2~\cite{ldp}  & 55.1 & \underline{86.8} & 76.1 & \textbf{71.4} & 30.1 & 60.1 & 1414 & \textbf{66.2} & 62.6 & \textbf{33.5} & \underline{98.61\%} \\
    \textbf{MS-Resampler} & \textbf{56.6} & \textbf{87.6} & \textbf{78.0} & \underline{70.6} & \textbf{30.8} & \textbf{62.1} & \textbf{1469} & \underline{64.5} & \textbf{66.1} & 32.9 & \textbf{100.07\%}\\
    \hline

    \rowcolor{gray!10}
    \multicolumn{12}{c}{ LLaVA-1.5-7B, Vision token = 36}\\
    \hline
    Pixel-Shuffle~\cite{InternVL15} & 48.1 & 84.6 & 70.6 & 68.7 & 25.8 & 57.2 & 1309 & 61.3 & 56.1 & \textbf{32.9} & 91.60\% \\
    FasterVLM~\cite{FasterVLM} & 53.7 & 76.6 & 71.5 & 68.8 & 27.5 & 52.5 & 1262 & 59.7 & 54.4 & 29.6 & 89.69\% \\
    C-Abstractor~\cite{cabs} & 49.6 & 85.7 & 72.4 & 69.0 & 29.3 & 58.3 & 1347 & 62.4 & 58.4 & 31.8 & 93.97\% \\
    MQT-LLaVA~\cite{MQT} & 51.0 & 81.9 & 75.3 & 66.8 & 27.8 & 60.0 & \textbf{1464} & 63.4 & 59.0 & 32.1 & 94.73\% \\
    TokenPacker~\cite{tp} & \underline{55.2} & \underline{86.5} & 76.1 & 67.8 & 30.2 & \underline{60.5} & 1420 & 63.5 & \underline{62.5} & 32.3 & 97.39\% \\
    SAEP~\cite{SAEP} & 51.7 & 85.4 & \underline{76.3} & 67.6 & 30.9 & 59.3 & 1356 & 60.5 & 61.7 & 31.1 & 95.30\% \\
    LDP-v2~\cite{ldp} & 53.6 & 86.4 & 75.2 & \underline{69.7} & \underline{31.7} & 59.9 & \underline{1443} & \underline{64.8} & 61.0 & 32.3 & \underline{97.77\%} \\
    \textbf{MS-Resampler}  & \textbf{55.8} & \textbf{87.2} & \textbf{76.8} & \textbf{69.9} & \textbf{32.8} & \textbf{61.2} & 1429 & \textbf{64.9} & \textbf{62.8} & \underline{32.6} & \textbf{99.32\%}\\
    \hline

    \rowcolor{gray!10}
    \multicolumn{12}{c}{ LLaVA-1.5-13B, Vision token = 144}\\
    \hline
    Vanilla & 63.8 & 87.9 & 80.0 & 71.6 & 35.4 & 63.3 & 1531 & 67.7 & 68.2 & 33.1 & 100.00\% \\
    Pixel-Shuffle~\cite{InternVL15} & 57.0 & 86.9 & 76.9 & 71.3 & 33.1 & 61.2 & 1483 & 66.2 & 64.0 & 32.1 & 95.96\% \\
    FasterVLM~\cite{FasterVLM} & \textbf{58.8} & 86.1 & 77.1 & 71.9 & 33.6 & 59.1 & 1489 & 66.4 & 63.8 & 32.3 & 96.17\%\\
    C-Abstractor~\cite{cabs} & 53.9 & 86.3 & 74.8 & 70.4 & 28.2 & 60.4 & 1391 & 63.0 & 61.0 & 31.3 & 91.75\%\\
    TokenPacker~\cite{tp} & \textbf{58.8} & 87.4 & \textbf{78.8} & 70.7 & 34.5 & \underline{62.3} & 1525 & \textbf{67.4} & \underline{65.4} & \textbf{33.9} & \underline{98.22\%}\\
    LDP-v2~\cite{ldp}   & 58.4 & \underline{87.6} & \underline{78.4} & \underline{72.3} & 34.4 & 62.0 & \underline{1533} & \underline{66.9} & 65.3 & 33.2 & 98.03\%\\
    \textbf{MS-Resampler}& \underline{58.5} & \textbf{87.9} & 78.3 & \textbf{72.8} & \underline{35.1} & \textbf{62.4} & \textbf{1547} & \underline{66.9} & \textbf{66.2} & \underline{33.7} & \textbf{98.77\%}\\
    \hline\hline
  \end{tabular}
  }
  \label{tab-1}
\end{table*}
\section{Experiment}
\subsection{Experimental Setup}

\paragraph{Model Architecture.}
We adopt a standard multimodal large language model (MLLM) architecture consisting of a vision encoder, a visual projector, and a large language model (LLM).
The vision encoder is CLIP-ViT-L/14~\cite{clip} with an input resolution of $336\times336$.
To evaluate the generality of MS-Resampler across different language backbones, we pair the same vision encoder with four LLMs of different architectures and scales: Vicuna-7B, Vicuna-13B, and Qwen2.5-3B~\cite{vicuna,Qwen25}.
Unless otherwise specified, MS-Resampler uses $E{=}3$ scoped resampling branches with window sizes $\{k^{(1)},k^{(2)},k^{(3)}\}=\{2,6,24\}$.

\paragraph{Training Data.}
We follow the data configuration of LLaVA-1.5~\cite{llava15}.
Specifically, we use the 558K image-caption pairs for pretraining the visual projector and the 665K multimodal instruction-following data for instruction tuning.
Unless explicitly stated, all training experiments are conducted using the LLaVA-1.5 training data to ensure fair comparison with existing methods.

\paragraph{Training Recipe.}
\label{sec:exp_setup}
All models are trained using 8 NVIDIA A800 GPUs with 80GB memory.
We adopt a two-stage training strategy.
In the first stage, only the visual projector is trained while freezing both the vision encoder and the LLM.
This stage uses the caption dataset, with a learning rate of $1\mathrm{e}{-3}$, batch size of 32, and is trained for one epoch.
In the second stage, we jointly train the visual projector and the LLM while keeping the vision encoder frozen.
This stage uses the instruction tuning dataset, with a learning rate of $1\mathrm{e}{-5}$, batch size of 16, and is trained for one epoch.
We use the AdamW optimizer for all experiments.

\subsection{Results on Vicuna-based MLLMs}
\paragraph{Impact of Spatial Attention Bias.}
We study the role of the spatial attention bias $\mathbf{B}^{(e)}$ in Scoped Resampler by varying the strength of the scope prior controlled by $\beta$ in Eq.~(\ref{eq:scoped_attn}). Specifically, we compare the full model against variants that remove the bias (setting $\beta=0$) or weaken/strengthen its effect by adjusting $\beta$, while keeping the number of branches, kernel sizes, and the token budget fixed.

Overall, introducing the spatial bias consistently improves performance over the $\beta=0$ variant, confirming that an explicit scope prior helps each branch specialize to its intended aggregation range. When $\beta$ is too small, the scoped branches behave similarly to global resampling and the benefit of multi-scope design diminishes. When $\beta$ is too large, the prior becomes overly restrictive and may hinder cross-scope aggregation, leading to degraded performance. These results suggest that a moderate bias strength provides the best trade-off: it encourages scope-aware aggregation while preserving the flexibility to attend beyond the preferred neighborhood when necessary.
\begin{table*}[t!]
  \centering
  \caption{Performance comparison with intra-LLM visual token compression methods on LLaVA-1.5-7B.}
  \setlength{\tabcolsep}{1.5pt} 
  \fontsize{8pt}{10pt}\selectfont
  \begin{tabular}{l|c | c c c c c c c | c}
    \hline\hline
     \multirow{2}{*}{Method} & Vision  & \multirow{2}{*}{VQA\textsuperscript{T}} & \multirow{2}{*}{POPE} & \multirow{2}{*}{VQA\textsuperscript{v2}}& \multirow{2}{*}{GQA}& \multirow{2}{*}{MMB} & MMB- & OCR & \multirow{2}{*}{Avg.}  \\
     & Token &  & & & & & CN& Bench\\
    \hline
    LLaVA-1.5-7B & 576 & 58.2 & 85.9 & 78.5 & 62.0 & 64.3 & 58.3 & 297 & 100.00\% \\
    \hline
    FastV~\cite{FastV} & 192 & 52.5 & 64.8 & 67.1 & 52.7 & 61.2 & 57.0 & 291 & 89.58\% \\
    HiRED~\cite{HiRED} & 192 & 47.4 & 82.8 & 74.9 & 58.7 & 62.8 & 54.7 & 190 & 89.06\% \\
    FitPrune~\cite{FitPrune} & 192 & 57.4 & 83.4 & -- & \underline{60.4} & 63.3 & 56.4 & -- & -- \\
    LLaVA-PruMerge~\cite{LLaVAprumerge} & 192 & 54.3 & 71.3 & 70.6 & 54.3 & 59.6 & 52.9 & 253 & 88.92\% \\
    SparseVLM~\cite{SparseVLM} & 192 & 56.1 & \underline{83.6} & 75.6 & 57.6 & 62.5 & 53.7 & 292 & 95.79\% \\
    PDrop~\cite{PDrop} & 192 & 56.1 & 82.3 & 75.1 & 57.1 & 63.2 & 56.8 & 290 & 96.19\% \\
    MustDrop~\cite{MustDrop} & 192 & 56.5 & 82.6 & 76.0 & 58.2 & 62.3 & 55.8 & 289 & 96.26\% \\
    DART~\cite{DART} & 192 & \textbf{57.4} & 82.8 & \underline{76.7} & 60.0 & \underline{63.6} & \underline{57.0} & \underline{296} & \underline{97.98\%} \\
    \textbf{MS-Resampler} & 144 & \underline{57.2} & \textbf{87.0} & \textbf{78.0} & \textbf{62.0} & \textbf{65.6} & \textbf{58.7} & \textbf{301} & \textbf{100.43\%}\\
    \hline\hline
  \end{tabular}
  \label{tab-2}
\end{table*}

\paragraph{Comparison with pre-LLM methods.}
We compare MS-Resampler with representative pre-LLM projectors that compress or transform visual tokens before entering the LLM, including the standard MLP projector and recent projector-level methods (FasterVLM~\cite{FasterVLM}, MQT-LLaVA~\cite{MQT}, Pixel-Shuffle~\cite{InternVL15}, C-Abstractor~\cite{cabs}, TokenPacker~\cite{tp}, LDP-v2~\cite{ldp}, and SAEP~\cite{SAEP}). 
Table~\ref{tab-1} reports results on Vicuna-7B and Vicuna-13B. 
With 144 tokens, MS-Resampler achieves the best overall performance on Vicuna-7B and remains top-ranked across most benchmarks, indicating consistent gains in both fine-grained perception and multimodal reasoning. 
As the token budget shrinks to 64 and 36, the gap to prior methods widens; MS-Resampler remains the strongest under aggressive compression, demonstrating robust information preservation with limited visual capacity. 
On Vicuna-13B (144 tokens), MS-Resampler again delivers the best average performance, suggesting that the proposed projector generalizes well across LLM scales.

\paragraph{Comparison with intra-LLM visual token compression methods.}
Table~\ref{tab-2} compares MS-Resampler with representative intra-LLM visual token compression methods on LLaVA-1.5-7B. 
Most of these baselines are training-free and reduce computation by pruning or merging visual tokens inside the LLM at inference time. 
To make the comparison conservative, we report MS-Resampler with 144 visual tokens, while using the default 192-token setting for training-free baselines.

Overall, MS-Resampler achieves the strongest average performance and slightly surpasses the full-token baseline, even under a smaller token budget. 
In contrast, intra-LLM compression methods consistently incur noticeable accuracy drops when reducing the token sequence, indicating that token pruning alone often sacrifices semantic coverage, especially for tasks requiring robust perception and fine-grained evidence. 
These results highlight a complementary perspective: improving the quality of compact visual representations at the projector stage can be more effective than relying solely on inference-time token reduction inside the LLM.

\begin{table*}[t!]
  \centering
  \caption{Performance comparison on Qwen2.5-3B.}
  \setlength{\tabcolsep}{2pt} 
  {
  \fontsize{8pt}{10pt}\selectfont
  \begin{tabular}{l | c | c c c c c c c | c}
    \hline\hline
    \multirow{2}{*}{Methods} & Vision & \multirow{2}{*}{VQA\textsuperscript{T}} & \multirow{2}{*}{POPE} & \multirow{2}{*}{VQA\textsuperscript{v2}} & \multirow{2}{*}{SQA} & \multirow{2}{*}{MM-Vet} & \multirow{2}{*}{GQA} & \multirow{2}{*}{MME\textsuperscript{P}} & \multirow{2}{*}{Avg.}\\
    & Token & & & & & & & & \\
    \hline
    \rowcolor{gray!10}
    \multicolumn{10}{c}{ Qwen2.5-3B, clip-vit-large-patch14-336} \\
    \hline
    Vanilla  & 576 & 54.7 & 85.9 & 76.9 & 72.3 & 31.2 & 59.8 & 1446 & 100.00\% \\
    Pixel-Shuffle~\cite{InternVL15} & 144 & 51.3 & 85.0 & 75.2 & 72.1 & 28.5 & 59.1 & 1379 & 96.54\% \\
    C-Abstractor~\cite{cabs}  & 144 & 48.0 & 84.6 & 73.6 & \textbf{72.8} & 29.4 & 57.8 & 1376 & 95.53\% \\
    LDP-v2~\cite{ldp} & 144 & 51.1 & 84.8 & 74.9 & 71.6 & 29.8 & 58.6 & 1394 & 96.93\% \\
    TokenPacker~\cite{tp} & 144 & 51.7 & 85.1 & 75.8 & 70.3 & \textbf{30.0} & 59.2 & 1388 & 97.23\% \\
    \textbf{MS-Resampler} & 144 & \textbf{52.4} & \textbf{85.4} & \textbf{76.2} & 71.3 & 29.4 & \textbf{60.0} & \textbf{1396} & \textbf{97.74\%} \\
    \hline
  \end{tabular}
  }
  \label{tab_3}
\end{table*}
\subsection{Generalization to Other LLM Backbones}
To verify that the gains of MS-Resampler are not specific to the Vicuna family, we further evaluate it on Qwen2.5-3B~\cite{Qwen25}. For a controlled comparison, we keep the vision encoder (CLIP-ViT-L/14@336), training data, and the two-stage training recipe identical to Sec.~\ref{sec:exp_setup}. We replace only the visual projector and keep the rest of the model unchanged.

Table~\ref{tab_3} summarizes the results. Under the same 144-token budget, MS-Resampler achieves the best overall average score among all projector-level compression baselines, improving over strong region-aware methods such as TokenPacker and LDP-v2. Beyond the aggregate metric, MS-Resampler remains competitive across diverse benchmarks that emphasize different capabilities, including knowledge-intensive and reasoning-oriented evaluations (e.g., SQA and MME) as well as general VQA benchmarks. Overall, the consistent improvements on Qwen2.5-3B indicate that the proposed multi-scope resampling is not tied to a particular language backbone, and can serve as a plug-and-play projector for different MLLM architectures and model scales.
\subsection{Ablation Studies}
We conduct ablation studies to understand the contributions of different components in MS-Resampler. 
Unless otherwise specified, all experiments are conducted on the Vicuna-7B backbone with the visual token budget fixed to 144. 
We follow the same training data and training recipe described in Sec.~\ref{sec:exp_setup}, and modify only the component under study while keeping the remaining settings unchanged.
\begin{table*}[t!]
  \centering
  \caption{Ablation study of the multi-scope design on LLaVA-1.5-7B.}
  \setlength{\tabcolsep}{1.5pt} 
  {
  \fontsize{8pt}{10pt}\selectfont
  \begin{tabular}{l | c c c c c c c c | c}
    \hline\hline
    Methods & VQA\textsuperscript{T} & POPE & VQA\textsuperscript{v2} & SQA & 
    GQA & MME\textsuperscript{P} & Seed\textsuperscript{I} & MMStar & Avg.\\
    \hline
    Vanilla  & 58.2 & 85.9 & 78.5 & 68.0 & 62.0 & 1511 & 66.1 & 31.9 & 100.00\%\\
    $E=1,k=2$  & 57.1 & \textbf{87.8} & 77.8 & 69.2 & 61.4 & 1451 & \textbf{65.2} & 32.4 & 99.56\%\\
    $E=1,k=6$  & 56.8 & 87.1 & 77.7 & 70.0 & 61.9 & 1452 & 64.8 & 32.9 & 99.75\% \\
    $E=1,k=8$  & 56.9 & 87.1 & 77.5 & 69.1 & 61.9 & 1463 & 64.8 & 32.8 & 99.63\%\\
    $E=1,k=12$ & 56.9 & 87.5 & 77.7 & 70.0 & 61.8 & 1464 & 64.9 & 33.1 & 100.01\%\\
    $E=1,k=24$ & 56.5 & 86.7 & 77.1 & 69.7 & 60.9 & \textbf{1475} & 64.3 & 33.7 & 99.69\%\\
    $E=3,k^{\{e\}}=\{2,6,14\}$ & \textbf{57.2} & 87.0 & \textbf{78.0} & \textbf{71.2} & \textbf{62.0} & 1453 & \textbf{65.2} & \textbf{33.9} & \textbf{100.59\%}\\
    \hline\hline
  \end{tabular}
  }
  \label{tab_4}
\end{table*}

\paragraph{Effect of Multi-Scope Design.}
We first examine whether multi-scope resampling is necessary beyond a single scoped branch. 
We collapse MS-Resampler into a single-branch variant ($E{=}1$) and vary the window size $k\in\{2,6,8,12,24\}$, while keeping the projected token budget fixed at $M{=}144$ and all other settings unchanged. Table~\ref{tab_4} reports the results.

Overall, the $E{=}1$ variants show a clear dependence on the chosen scope: smaller windows tend to favor localized evidence (e.g., stronger TextVQA), whereas larger windows better capture broader context (e.g., stronger MME/MMStar), indicating an inherent trade-off under a single aggregation range. No single $k$ consistently performs best across benchmarks, and the best single-branch setting remains below the multi-scope model on average. In contrast, MS-Resampler with $E{=}3$ and $k^{\{e\}}=\{2,6,14\}$ achieves the strongest overall performance in Table~\ref{tab_4}, outperforming all single-branch counterparts. This improvement suggests that combining complementary scopes is critical for preserving both fine-grained evidence and global context under a fixed token budget.

\begin{table*}[t!]
  \centering
  \caption{Ablation study of kernel size configurations in MS-Resampler on LLaVA-1.5-7B ($E{=}3$, $M{=}144$).}
  \setlength{\tabcolsep}{1.5pt} 
  {
  \fontsize{8}{10}\selectfont
  \begin{tabular}{l | c c c c c c c c | c}
    \hline\hline
    Methods & VQA\textsuperscript{T} & POPE & VQA\textsuperscript{v2} & SQA & 
    GQA & MME\textsuperscript{P} & Seed\textsuperscript{I} & MMStar & Avg.\\
    \hline
    Vanilla  & 58.2 & 85.9 & 78.5 & 68.0 & 62.0 & 1511 & 66.1 & 31.9 & 100.00\%\\
    $k^{\{e\}}=\{2,3,5\}$   & 57.0 & \textbf{87.3} & 77.4 & 69.4 & 61.6 & 1451 & 65.0 & 32.5 & 99.48\%\\
    $k^{\{e\}}=\{6,8,12\}$  & 55.7 & 86.9 & 77.8 & 69.2 & 60.9 & 1412 & 63.57 & 31.9 & 98.20\% \\
    $k^{\{e\}}=\{15,18,24\}$& 56.5 & 86.7 & 77.3 & 69.8 & 61.3 & \textbf{1475} & \textbf{65.2} & 33.1 & 99.68\%\\
    $k^{\{e\}}=\{2,6,14\}$  & \textbf{57.2} & 87.0 & \textbf{78.0} & \textbf{71.2} & \textbf{62.0} & 1453 & \textbf{65.2} & \textbf{33.9} & \textbf{100.59\%}\\
    \hline\hline
  \end{tabular}
  }
  \label{tab_5}
\end{table*}

\paragraph{Effect of Kernel Size.}
Table~\ref{tab_5} studies the choice of scope sizes in MS-Resampler with $E{=}3$ and $M{=}144$. We evaluate four kernel sets: $k^{\{e\}}=\{2,3,5\}$, $\{6,8,12\}$, $\{15,18,24\}$, and $\{2,6,14\}$. Configurations concentrated on small kernels tend to emphasize local evidence (e.g., stronger POPE), while large kernels behave closer to global resampling and reduce scope diversity. Among all settings, the mixed configuration $\{2,6,14\}$, spanning local-to-large scopes, achieves the best overall average and consistently strong results, suggesting that covering a broad range of spatial granularities is important for effective multi-scope resampling.

\paragraph{Fusion Strategy.}
We ablate the fusion module while keeping the multi-scope branches and scope priors fixed. Let the output of branch $e$ at token position $i$ be $\mathbf{o}^{(e)}_{i}\in\mathbb{R}^{c}$. All fusion variants are \emph{position-wise}: for each token index $i$, we fuse $\{\mathbf{o}^{(e)}_{i}\}_{e=1}^{E}$ into $\mathbf{o}^{\mathrm{fuse}}_{i}\in\mathbb{R}^{c}$ without mixing representations across different token positions, ensuring that any performance difference is attributable to the fusion design. We compare four fusion strategies in Table~\ref{tab_6}. \textbf{Average} fusion performs element-wise mean pooling across branches, i.e., $\mathbf{o}^{\mathrm{fuse}}_{i}=\frac{1}{E}\sum_{e=1}^{E}\mathbf{o}^{(e)}_{i}$, which is stable but cannot adapt the contribution of different scopes. \textbf{Attention} fusion learns scope weights at each token position and forms a weighted sum, $\mathbf{o}^{\mathrm{fuse}}_{i}=\sum_{e=1}^{E}\alpha^{(e)}_{i}\mathbf{o}^{(e)}_{i}$, where $\{\alpha^{(e)}_{i}\}$ are produced by a lightweight attention over the $E$ branch features. \textbf{MLP} (channel-mixed) fusion concatenates branch outputs along the feature dimension, $\mathbf{u}_i=[\mathbf{o}^{(1)}_{i};\dots;\mathbf{o}^{(E)}_{i}]\in\mathbb{R}^{Ec}$, and projects it back to $c$ dimensions with an MLP, allowing joint mixing across scopes and channels. \textbf{MS-Resampler} uses our channel-wise MLP fusion: for each channel $j$, we form $\mathbf{z}_{i,j}=[o^{(1)}_{i,j},\dots,o^{(E)}_{i,j}]^\top\in\mathbb{R}^{E}$ and apply a shared two-layer MLP along the scope dimension to obtain $o^{\mathrm{fuse}}_{i,j}$, which preserves token alignment and performs lightweight scope arbitration per channel.

Table~\ref{tab_6} shows that simple averaging is a strong baseline but remains below our default fusion. Attention fusion brings limited gains and is less consistent across benchmarks. Channel-mixed MLP fusion performs worst, suggesting that aggressive cross-scope channel mixing can hurt compact visual representations. In contrast, the channel-wise fusion in MS-Resampler achieves the best overall average and consistently strong results, indicating that per-channel scope arbitration offers a better balance between expressiveness and stability under a fixed token budget.

\begin{table*}[t!]
  \centering
  \caption{Ablation study of fusion strategies in MS-Resampler on LLaVA-1.5-7B.}
  \setlength{\tabcolsep}{1.5pt} 
  {
  \fontsize{8}{10}\selectfont
  \begin{tabular}{l | c c c c c c c c | c}
    \hline\hline
    Methods & VQA\textsuperscript{T} & POPE & VQA\textsuperscript{v2} & SQA & 
    GQA & MME\textsuperscript{P} & Seed\textsuperscript{I} & MMStar & Avg.\\
    \hline
    Vanilla  & 58.2 & 85.9 & 78.5 & 68.0 & 62.0 & 1511 & 66.1 & 31.9 & 100.00\%\\
    Average  & 57.1 & \textbf{87.3} & 77.6 & 70.1 & 61.9 & \textbf{1469} & 65.1 & 31.5 & 99.50\% \\
    Attention& 55.5 & 87.2 & 77.5 & 70.5 & 61.8 & 1441 & 64.4 & 32.9 & 99.36\% \\
    MLP      & 55.2 & 86.8 & 77.1 & 70.2 & 61.1 & 1399 & 63.8 & 33.2 & 98.63\% \\
    MS-Resampler & \textbf{57.2} & 87.0 & \textbf{78.0} & \textbf{71.2} & \textbf{62.0} & 1453 & \textbf{65.2} & \textbf{33.9} & \textbf{100.59\%}\\
    \hline\hline
  \end{tabular}
  }
  \label{tab_6}
\end{table*}

\begin{table}[b!]
    \setlength{\tabcolsep}{1mm} 
    \centering
    \caption{Complexity comparison between QMoP and baseline (LLaVA-1.5-7B).}
     {
    \fontsize{8}{10}\selectfont 
    \begin{tabular}{l| c |c c c c c c}
    \hline
    \hline
    \multirow{2}{*}{Methods} & Vision & Total & Projector & \multirow{2}{*}{KVcache} & Train & Inference  & \multirow{2}{*}{Metric} \\
    & Token & FLOPS &FLOPS &  & time & time & \\
    \hline
    LLaVA-1.5-7B & 576 & 3.82 T & 12.09 G & 302.0 M & 102.3 h & 3h05m & 100\%\\
    \hline
    C-Abstractor & 144 & 0.94 T & 12.09 G & 75.5 M  & 72.3 h  & 2h22m & 95.91\%\\
    TokenPacker & 144 & 0.94 T & 10.73 G & 75.5 M  & 74.2 h  & 2h24m & 100.25\%\\
    LDP-v2 & 144 & 0.94 T & 12.10 G & 75.5 M  & 73.1 h  & 2h21m & 99.25\%\\
    MS-Resampler & 144 & 0.94 T& 10.43 G & 75.5 M   & 73.5 h  & 2h18m & 101.15\%\\
    \hline
    \hline
    \end{tabular}
    }
    \label{tab_8}
\end{table}
\subsection{Efficiency Analysis}
Table~\ref{tab_8} reports a complexity comparison between MS-Resampler and representative projector-level compression baselines on LLaVA-1.5-7B. By reducing the visual token length, all compressed projectors substantially lower the overall computation and KV cache footprint compared to the 576-token vanilla model, leading to faster training and inference. 
Within this compressed regime, MS-Resampler remains efficient. It matches the total FLOPs and KV cache of other 144-token methods, while incurring no extra memory overhead. Meanwhile, MS-Resampler has a lightweight projector with comparable (or lower) FLOPs than prior projector designs, and achieves the fastest inference time among the compared methods. Importantly, these efficiency gains do not come at the cost of accuracy: MS-Resampler delivers the best overall performance in Table~\ref{tab_8}. Overall, the results show that multi-scope resampling provides a favorable trade-off, improving multimodal performance while maintaining the efficiency benefits of aggressive visual token reduction.
\section{Conclusion}
We proposed MS-Resampler, a multi-scope visual projector for multimodal LLMs. It combines multiple scope-controlled resampling branches and a lightweight fusion module to encode local-to-global visual semantics under a fixed token budget. Experiments on diverse benchmarks and multiple LLM backbones show consistent gains over strong projector-level baselines and competitive performance against intra-LLM token compression methods, especially under aggressive token reduction.


%
%
\bibliographystyle{splncs04}
\bibliography{main}
\end{document}